# Design Strategy Network: A deep hierarchical framework to represent generative design strategies in complex action spaces

**Ayush Raina**
Department of Mechanical Engineering
Carnegie Mellon University
Pittsburgh, PA, USA
araina@andrew.cmu.edu

**Jonathan Cagan**
Department of Mechanical Engineering
Carnegie Mellon University
Pittsburgh, PA, USA
cagan@cmu.edu

**Christopher McComb**
Department of Mechanical Engineering
Carnegie Mellon University
Pittsburgh, PA, USA
ccm@cmu.edu

**ABSTRACT**
*Generative design problems often encompass complex action spaces that may be divergent over time, contain state-dependent constraints, or involve hybrid (discrete and continuous) domains. To address those challenges, this work introduces Design Strategy Network (DSN), a data-driven deep hierarchical framework that can learn strategies over these arbitrary complex action spaces. The hierarchical architecture decomposes every action decision into first predicting a preferred spatial region in the design space and then outputting a probability distribution over a set of possible actions from that region. This framework comprises a convolutional encoder to work with image-based design state representations, a multi-layer perceptron to predict a spatial region, and a weight-sharing network to generate a probability distribution over unordered set-based inputs of feasible actions. Applied to a truss design study, the framework learns to predict the actions of human designers in the study, capturing their truss generation strategies in the process. Results show that DSNs significantly outperform non-hierarchical methods of policy representation, demonstrating their superiority in complex action space problems.*

## 1. INTRODUCTION

The process used to create a product is often as important as the finished product itself. This is especially true in generative design problems that require iterative interaction with a concept before creating a finished product [1,2]. Designers take numerous sequential decisions to search and explore large design spaces, eventually creating products that satisfy their objectives. Expert designers leverage their domain knowledge and learned heuristics as intuition to filter the design space and reach solutions quicker [3,4]. For any given design problem, numerous decision-making strategies exist that lead to a variety of solutions. However, only some of these strategies consistently lead to high-performing solutions, and being able to extract and model such effective strategies is a critical step towards building intelligent agents. This work introduces a framework that extends design strategy representation to a generic action space while implicitly modeling the process's underlying relationships.

This research focuses on parameterized configuration design problems [5] modeled as sequential generative design problems involving numerous decision-making steps [6,7]. These problems require generating a configuration from a set of parameterized components in incremental steps such that the overall configuration satisfies an objective. These decisions involve placing the components at different locations and adjusting their parameters. Hence, both the content (the *what* and the *where*) and context (the *when*) of decisions is essential to model the decision-making process. Content is associated with the action definition, which often includes spatial parameters as continuous variables, action types as discrete variables, and feasibility based on the several constraints of the problem. Hence, a representation that captures this constrained hybrid (discrete and continuous) information is essential. Context captures the relationship between design states and selected action since a given action may be beneficial in a particular state but detrimental somewhere else. Further, as the design progresses and becomes more complex in a configuration design process, the number of components and their associated actions also increase; this is referred to as a divergent tree expansion where the feasible action space expands over time. Other problem-solving processes like chess, Go, and many other board games have characteristically convergent tree expansions since the number of pieces and possible spaces one can move to decrease with the game's progress [8]. As a result, a method that models the decision-making process of sequentially generative design problems must address the challenges associated with a divergent and constrained hybrid action space. This work specifically addresses this problem of arbitrarily complex action spaces through a uniquely designed, hierarchical data-driven architecture that mitigates the need for problem-specific action inference methods.

Modeling decision making in problem-solving processes can be formally defined as learning a function $\pi$ that represents the strategy (or policy) followed by the solver, where $\pi_\theta(s_t) = Prob(A_t)$ [9]. This function uses the state information ($s_t$) as input and outputs a probability distribution, $Prob(.)$, over the set of possible actions ($A_t$) to make the final decision at a particular time $t$ in the process [2,9]. Here $\theta$ is the set of parameters associated with the function model that need to be learned. Existing Reinforcement Learning and Imitation Learning methods use deep networks to learn this policy function $\pi$ from largescale data of humans or semi-supervised training to learn it with minimal data [9,11–13]. For specific problems, these deep policy networks can learn both optimal action selection and which actions are feasible and infeasible based on environment dynamics [14,15]. However, this success is limited to applications with small action spaces, require large datasets with millions of data points, and often show improper generalization to novel states [16]. Several works in the engineering design area have made progress in capturing human strategies [17–22]. However, they limit the definition of actions to high-level commands or stages in a design process to simplify the learning complexity, a framing not sufficient for the learning problem at hand.

This paper surpasses the detail level in previous work by learning the relationships between an image-based representation of the design state and a comprehensive action representation by using a novel hierarchical approach to decompose the action prediction step. The comprehensive action representation includes all parameters associated with the action definition. This research goal of the current work centers around leveraging deep learning tools to extract and represent design strategies and, by extension, design knowledge purely from data. The main research question the proposed methodology aims to answer is: *Can data-driven learning with hierarchical actions emulate human decision-making behavior in complex action spaces better than a baseline non-hierarchical method?*

To explore this question, a novel hierarchical architecture termed Design Strategy Network (DSN) is proposed. This framework can capture and represent strategies in divergent and constrained hybrid action spaces. It breaks down the prediction of the final action into first predicting a rough spatial region in the design space. This breakdown is analogous to a human designer's visual attention, which helps them focus on only the critical regions of a design space. Once the spatial region is identified, a set of possible actions in that region are queried from the design environment, which is then utilized by the framework to generate a probability distribution over actions. This hierarchy enables the framework to learn an explicit relationship between the state and the final detailed action information. Since the framework queries the set of feasible actions directly from the design environment, it need not learn to identify feasible actions in arbitrary states. Instead, the DSN focuses on representing the relationships between the design state and individual actions from a variable set of feasible actions. The DSN framework works comprehensively in a data-driven manner and does not require problem-specific inference algorithms to make the action decisions, allowing the learning process to be generic and problem-independent. This work explores a case study of a truss design problem [23], which models the classic truss problem in engineering design [24] as a sequential decision-making process. The truss design problem is formulated as a unique Markov Decision Process with divergent and constrained hybrid action spaces as previously described in [25]. This truss design problem captures various complexities of a design problem such as high-dimensionality, image-based data, implicit physical dynamics in state representations, no preliminary design feedback, and a complex action space with constrained continuous and discrete parameters. Hence, it provides a challenging test for evaluating DSNs and illustrates how the framework adapts to the different challenges.

The remaining paper is divided into five sections. Section 2 provides the relevant literature for the various components of the proposed methodology. Section 3 presents the method introducing the hierarchical DSN architecture and its training pipeline. Section 4 details the truss design case study used in this work, along with the network details. Section 5 presents the validation results and relevant discussion of the framework performance. Lastly, Section 6 concludes the learnings from the validation of the proposed framework.

## 2. Background

The background section broadly covers the relevant literature of the concepts utilized in this work divided into three sub-sections: The first subsection covers similarly motivated works. The second covers other frameworks that use a similar methodology of hierarchical prediction in imitation learning. The third subsection captures a review of different components used in the architecture of the framework.

### 2.1 Modeling decision making in design problem solving

Engineering design is often considered as an elaborate sequence of decisions [2]. Formulating design in that manner allows proper formalisms in terms of the state's definitions, the possible actions, the relevant objectives, and constraints. Decision-Based Design is a perspective that builds upon this formulation. It acknowledges decision-making challenges regarding ambiguity, tradeoff, and uncertainty in the design process [7,26,27]. These ideas allow the application of mathematical analyses and tools to guide the process. Modeling design decisions have been achieved with Bayesian models, where models extract frequently occurring design sequences as probability distributions with sequential actions. Frequently occurring design actions have been extracted using first and second-order Markov Chain models [17,28,29]. Hidden Markov Models have also been shown to learn a discriminative representation between high and low skilled strategies in human designers [18]. Further, these representations were seen to maintain their characteristics when transferred to design agents or new problems [30], enabling transfer learning. Apart from Bayesian methods, Gaussian processes have also been used to capture human decision-making behavior and have been used for modeling search [31], choosing design parameters [31], and even making decisions on information acquisition [32,33]. Other recent work has also employed deep learning models to predict human action decisions and capture their behavior. Rahman et al. [19] illustrate that combining static and dynamic information enables better prediction using a recurrent neural network. Bayrak and Sha [21] integrate game theory and sequence learning using LSTM networks to predict design decisions. Raina et al. [34] introduce Deep Learning Agents (DLAgents) that illustrate the efficacy of using visual imitation-based methods to model human decision-making in design. These agents used a combination of data-driven strategies and image processing-based inference algorithms to simulate human decision-making. Our current work extends the DLAgents framework to replace the rule-based inference algorithms with a data-driven framework, extending its applicability to other similar problems with minimal additional effort.

### 2.2 Hierarchical policy network: Integrating spatial intuition to guide action selection

A common type of Reinforcement Learning algorithm, policy networks [9], usually implicitly define a simple action space [12]. Networks that deal with hybrid action spaces (discrete and continuous actions) adopt specialized architectures capable of accommodating the additional complexity [35–38]. Further, for specific problems, it is necessary to filter actions based on state constraints [39,40]. This mechanism works for discrete actions but cannot be extended to continuous action spaces. The combination of state constraints and hybrid action spaces has only been addressed in large-scale deep learning methods [15,41]; these methods depend on imitation learning over millions of data points to effectively learn the constraints from environment dynamics, enabling them to determine the feasible action space implicitly. The differences among the action spaces can be summarized as follows: Deep Q Networks (DQN) and its variants [42,43] are developed for discrete actions where every neuron output corresponds to an action. Deep Deterministic Policy Gradients (DDPG) [44] are used for continuous vector outputs as action values, where the losses are computed based on an approximated policy gradient formulation. Parameterized DDPG [37] and Parameterized DQN [45] methods are developed for hybrid actions spaces (having both discrete and continuous actions) with inverting gradient losses and custom architectures, however, these methods are only limited to those domains.

This paper addresses the challenge of action space representation by decomposing the task of action selection into a hierarchical two-step decision. The first step predicts a preferred spatial region in the state space, analogous to hard attention that filters out unimportant regions; such hard attention has been previously used for computer vision tasks [46–48]. Since hard attention explicitly outputs the desired spatial region, it allows effective interfacing with a non-differentiable design environment function for sampling feasible actions. Similar spatial intuition or visual attention-based approaches have recently emerged in Reinforcement Learning literature with great success and numerous applications [49–55]. Our work is similar to the framework of Mott et al., which used soft-attention-based RL agents [55]; however, their attention mechanism is parallel to the network, which is different from our hierarchical approach. The hierarchical approach first roughly predicts the spatial region before predicting the exact value of the final action. The presented DSN framework has hard spatial attention as an indispensable part of the policy network, which, along with the other parts of the framework, allows usage in constrained hybrid action spaces. This hierarchical approach enables sampling feasible actions from the design environment, and hence the method can be applied to a wide array of problems. Moreover, the preferred spatial region promotes a more interpretable decision-making approach since the spatial region it generates provides insights about the internal decision-making of the framework, which can be helpful in human-machine collaborative settings or a qualitative evaluation of the decision-making behavior.

## 2.3 Deep learning on image-based states and unordered action sets

Design processes often involve decision-making based on visual state information wherein human designers interact with image-based snapshots of design state representation. The high dimensionality of image-based data makes learning over it complicated, requiring a representation with reduced dimensions that preserves essential information from the original state. Recent advancements in machine learning have developed numerous methods to deal with such high-dimensional image-based data. Specifically, several deep network architectures that employ convolutional neural networks have been developed that effectively capture spatial features and generate a low-dimensional embedding [56,57]. Some recent works have introduced novel architectures from natural language processing for image-based data, even outperforming the existing methods [58,59]. However, for the scope of this research, a basic Atari Deep Q-Network [42] (Atari-DQN) based architecture is adopted due to its simplicity and its relatively low data requirements when compared with newer styles and deeper networks like VGG [56], ResNets [57] and Transformer networks [58].

Design problems often involve highly complex action representation. The action space can be variable depending on the state and include continuous parameters. Developing a deep learning network that can accommodate these characteristics is challenging because networks usually have a fixed-sized input. Such networks assign fixed weights to corresponding indexes giving importance to the order of the input. Complex action spaces require an order invariant set-based representation of action input data. Recently, there has been significant research in this area inspired by the point cloud representation of 3D models [60–62]. Studies have further enhanced these ideas' capabilities by introducing a deep-set formulation of neural networks that can accommodate such data's complexities [60,63–65]. Deep sets present the concept of invariant neural networks that transform each element of a set using multiple layers individually. This representation creates a transformation of the input set with each element independent of other elements. This representation is usually followed by some form of element aggregation and finally using a non-linear transformation to generate the relevant output. This method has been adopted in the presented work to capture the action space's complexities, treating the feasible actions as an unordered set input. Contrary to previously discussed methods of reinforcement learning (DQN, DDPG, Parameterized DQN and Parameterized DDPG), the proposed new framework with a dynamic feasible action input in every state can be applied to a complex action space that includes hybrid actions and state-dependent constraints as it can deal with a variable sized set input for every state. The proposed framework's selection network takes the unordered set of actions as input and combines them with state encoding before aggregating and generating a probability distribution over the action set.

## 3. DESIGN STRATEGY NETWORK (DSN)

This work introduces Design Strategy Network as a generic framework for representing strategies in a design problem. DSN is a data-driven method that trains on historical trajectory data and instantaneously generates an action probability distribution given a state input. The ability to emulate decision-making behavior is analogous to representing design knowledge about solving problems, or *intuition*, through a pixel-based state input. The framework uses a hierarchical architecture that first predicts a preferred spatial region in the design space and then generates a probability distribution over a variable set of feasible actions in that region. A combination of three networks in the architecture collectively enables handling the problem's action space variability. Figure 1 schematically represents the architecture of the DSN framework, with the arrows representing the flow of information during inference or prediction. Three networks compose the architecture: Encoder network, Spatial Action network, and Selection network. The Encoder network downsamples the state information into a low dimensional encoding which helps the Spatial Action network to predict a preferred spatial region. This spatial region (shaded area) is then used to query the environment to sample a set of feasible actions. Finally, the Selection network uses both state encoding and possible action set to generate a probability distribution for final action selection. This unique hierarchical formulation of the framework allows effective decomposition of the tasks and presents the following set of features as contributions:

- *Purely data-driven approach:* Unlike previous approaches [25,34,66], DSN does not require rule-based inference algorithms to make action selections and is a data-driven methodology to predict human actions.

- *Modularity:* The framework is decomposed into three different deep networks that have independent learning tasks. These networks can be independently improved upon and then merged into the overall framework.

- *Generic action space representation:* The framework is independent of the problem's action space type and can work with any arbitrary action space. Input to the framework is a variable-length set of feasible actions extending its applicability as a generic strategy representation framework.

- *Interpretable spatial intuition:* The framework predicts a rough region of importance in the design space as an intermediate step, making the internal decision-making more observable than an end-to-end method with no intermediate non-latent representations. Further, this predicted region is also used to interface with the environment to sample the feasible actions.

The following sub-sections detail the architecture (Section 3.1), sampling mechanism (Section 3.2), and training process (Section 3.3 and Section 3.4) for the hierarchically structured DSNs.

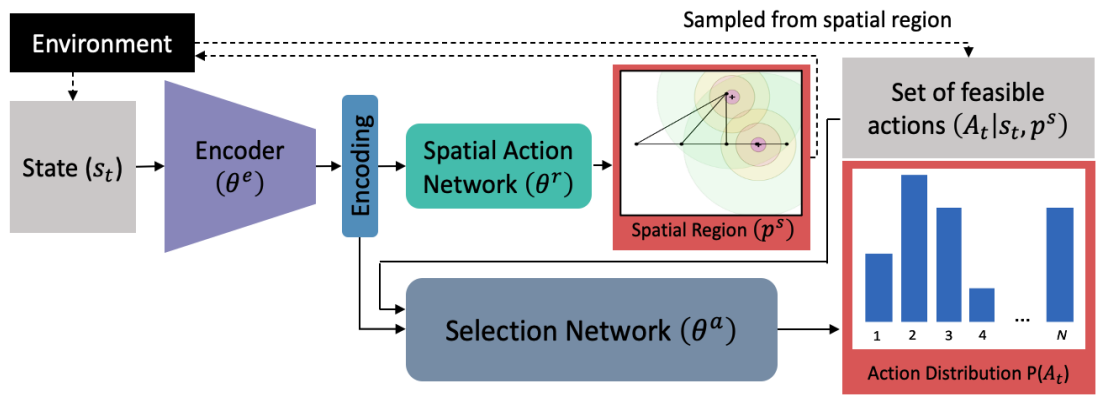

*Figure 1:* Schematic for DSN architecture. Solid lines represent the flow of information through differentiable components of the network. Dashed lines represent the queries to the design environment (non-differentiable).

## 3.1 Hierarchical architecture

The proposed framework decomposes the problem into three independent networks that feed into one another and make the final action selection. This decomposition's inspiration comes from the human visual attention mechanism, allowing humans to focus on only a tiny part of a problem [3,67]. Over numerous years of experience, expert humans instinctively identify promising regions of a solution space to focus their cognitive load effectively. Implementing this introduces a hierarchy in the architecture since the action prediction task is split into first predicting the "parent" spatial region, which governs a set of "children" feasible actions, one of which is then selected as the final action prediction. This breakdown emulates the expert behavior for quickly filtering the solution space while also enabling integration with the environment function to sample a set of continuous actions enabling the hierarchical network to interface with a complex action space.

The input is first passed through the Encoder network, which transforms the image-based design state $(s_t)$ into a low-dimensional latent state encoding $(s_t^*)$ while capturing its essential information, as shown in Eq. (1):

$$s_t^* = \mathbf{ENC}_{\theta^e}(s_t).\qquad(1)$$

Encoder network ($\boldsymbol{ENC}_{\theta^e}$ whose parameters are defined as $\theta^e$) comprises multiple convolutional layers that process the design state image and are employed to capture both low and high-level convolutional features. Next, the Spatial Action network ($\boldsymbol{SPA}_{\theta^r}$ with parameters $\theta^r$) uses the generated latent encoding to predict a spatial region corresponding to the spatial parameters($p^s$) of the final action. The Spatial Action network is a multi-layer perceptron that transforms the encoding into continuous spatial parameters, as shown in Eq. (2):

$$p^s = \mathbf{SPA}_{\theta^r}(s_t^*).\qquad(2)$$

The definition and dimensionality of the region and spatial parameters depend on the specific problem and must be uniquely identified for the use case. A spatial sampling method (defined in subsection 3.2) uses this predicted spatial region to query a feasible set of actions $(A_t \,|s_t, p^s)$ from the design environment. This process is represented as a dotted line in Figure 1 since it is not a part of the data-driven framework, but it is still controlled by the Spatial Action network's output as described in Eq. (3):

$$A_t = Env(s_t, p^s).\qquad(3)$$

Finally, the Selection network ($\boldsymbol{SEL}_{\theta^a}$ with parameters $\theta^a$) effectively evaluates all the actions in the unordered set of feasible actions and computes a probability distribution across the unordered set to select the appropriate action:

$$p^d = \mathbf{SEL}_{\theta^a}(s_t^*, A_t).\qquad(4)$$

The Selection network follows a Deep-Set formulation [63] to adapt to the unordered nature of the data. The architecture includes layers with shared weights that individually transform every action in the set and develops a state-action combined encoding before transforming into a probability distribution. This formulation is analogous to a Q-function since it takes both state and feasible actions as input and generates a probability equivalent to the expected value of the action in that state. Formulating the strategy network in this hierarchical format allows its application to any arbitrary action space definition.

Equations (1), (2), and (4) compose the data-driven part of the framework that focuses on learning the relationships between the selected action (with its associated parameters) and the design state information. Equation (3) generates a sample of feasible actions by querying the design environment, preventing the data-driven component's need to learn these relationships and hence, simplifying the learning problem. Since sampling feasible actions in a design state is not directly related to the decision-making mechanisms in a design problem, sampling it from the design environment allows the data-driven components to focus on learning to predict the problem-solving behavior.

## 3.2 Spatial sampling of feasible actions using the design environment

This subsection details a function to generate a tractable list of feasible actions in a given design state. In the case of design problems, the environment usually provides an interface to a user along with a list of feasible actions or tools to help them interact with the current design state. The underlying mechanism or physical models required to simulate the design are considered a black box and are not explicitly available

        

to the user. Ideally, the environment must determine an exhaustive list of all actions for the user; however, this number can be very high or infinite in the case of continuous action space. To circumvent this issue, a spatial sampling function Eq. (3) is developed that takes a preferred spatial region and generates a set of feasible actions. The spatial region is arbitrarily defined as a mixture of $S$ multivariate unit-variance (isotropic) Gaussian distributions. These $S$ different Gaussian distributions' dimensionality depends on the specific use case and can be modified arbitrarily given the problem and action definition. For instance, consider a problem in which every action requires three control points in x-y coordinates (illustrated in Figure 2). The relevant spatial region for such a problem is a mixture of 3 bivariate Gaussian distributions. In Figure 2, each distribution center is shown as '+', a rough probability distribution is shown as the shaded part, and the triangle represents the existing components in the region. Based on the spatial region, feasible actions are sampled and generated by this function. This function treats continuous parametric actions and state-dependent discrete actions differently. Since there is an infinite possible set of actions for continuous actions, only $n$ actions are sampled from this mixture of Gaussian distributions (shown as isolated circles in Figure 2). For discrete actions, a comprehensive list of actions based on the existing components present in the region is prepared, and then $A_{max} - n$ actions are selected based on their likelihood given the spatial region. As a result, a maximum of $A_{max}$ actions are sampled from the spatial region; $n$ continuous and $A_{max} - n$ discrete actions. These actions are computed based on feasibility and certain states may not contain $A_{max} - n$ feasible discrete actions, and hence the set of feasible action is variable. It must also be noted that this function does not provide any information about solving the problem and just aims to list the feasible actions in the preferred region and hence is a part of the design environment itself. The values of the introduced hyperparameters $(S, A_{max}, n)$ are defined with respect to the problem in subsection 4.1.

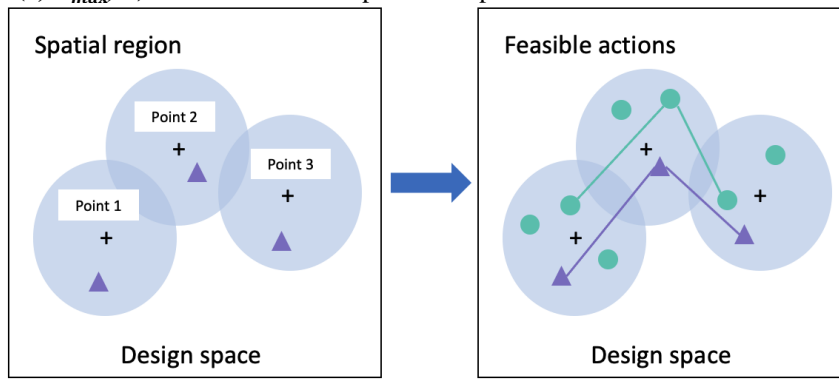

*Figure 2:* An example showing sampling feasible actions based on spatial region. Spatial regions for three control points are centered at respective '+'. Triangles show existing components. For feasible actions, isolated circles represent the continuous actions sampled in the region. The connected triangles and circles show discrete actions that require three control points.

### 3.3 Dataset processing

Training the hierarchical DSN framework requires a dataset with sequential decision-making information. The dataset must contain several design process trajectories, starting from initial design states and ending with the final design. The sequential dataset, D, needs to be of form D = $\{s_t, (p_t^d, p_t^s)\}$ $\forall$ t = [1, T] with T data samples, where $s_t$ is the design state represented as a multi-dimensional vector (pixels, voxel, or other matrix-based representations), $p_t^d$ is the discrete label of the selected action, and $p_t^s$ is the vector containing the spatial parameters of the selected action, at iteration $t$. Further, two different variants of this dataset must be formulated. The first dataset $D_1 = \{s_t; p_t^s\}$ $\forall t = [1, T]$ has the design state as the input and the continuous spatial parameters as the output. This dataset captures the intuitive relationship between the current state and the spatial region of importance based on where the human took the following action in the dataset. The other dataset $D_2 = \{(s_t, (p_t^d, p_t^s)^{A_{max}}); i\}$ $\forall i = [1, A_{max}]$ contains the current state and a set of top $A_{max}$ feasible actions as the input. The output is the classification index $i$ of the selected action. This dataset can be developed by recording human designers solving a problem or being synthetically generated by agents with high-performing behavior.

The dataset stores the associations between the combined set of states, feasible actions, and the final selected action capturing the domain knowledge integral to the sequential decision-making in the design process. The feasible actions are sampled using the methodology explained in Section 3.2 by utilizing the spatial region around the final selected action. Providing this set of feasible actions simplifies the learning problem for the learning framework since then the framework only needs to predict the best action from the list and not have to learn the relationship between the current state and the whole set of feasible actions. This sequential decision-making corresponds to the learning problem of one-step prediction and focuses on the immediate or myopic design strategies while long-term strategies are not explicitly modeled.

### 3.4 Training procedure

The DSN framework has two main flow hierarchies. The higher-level hierarchy outputs a spatial region $(p_t^s)$, and consists of the Encoder Network and Spatial Action network parameters. The lower-level hierarchy outputs a probability distribution over the set of feasible actions that includes both the spatial and discrete component of the action definition $(p_t^s, p_t^d)$ and consists of Encoder network and Selection network parameters. The two datasets $D_1$ and $D_2$, are required to train these two parts of the framework separately. Since the spatial region output is a continuous vector of spatial locations, a Mean Square Error (MSE) loss is used as in Equation (5):

$$min_{\theta^e, \theta^r}^{D_1} \left( p_{actual}^r - SPA_{\theta^r}(ENC_{\theta^e}(s_t)) \right)^2. \tag{5}$$

A Binary Cross-Entropy (BCE) loss trains the probability distribution output from the Selection network as shown in Equation (6):

$$min_{\theta^e, \theta^a}^{D_2} P(A_t)_{actual}. log(SEL_{\theta^a}(ENC_{\theta^e}(s_t), A_t)). \tag{6}$$

Both the losses are equally weighted during training. The MSE loss trains the network to predict the spatial parameters associated with the region, while the BCE loss compares the action distribution generated by the framework to one hot vector for the selected action index. The BCE loss trains the network to assign high value to the combined state-action encoding for the human selected action and lowers the values for all other options, effectively training the DSN framework to learn human action preferences given the state. It must be noted that encoder parameters $\theta^e$ are being optimized twice such that the essential features of the latent encoding include information for both spatial region prediction and the final action selection. The encoder network cannot be trained independently because it is a hidden representation, and there are no explicit labels available.

## 4. CASE STUDY: TRUSS DESIGN PROBLEM

The first subsection introduces the details of the truss design problem, the human design study conducted previously, and the preprocessing of the dataset. The second subsection provides the network details and other hyperparameters in the overall training setup. The third subsection provides the details of the experimental setup.

### 4.1 Problem definition and dataset preprocessing

The design of truss structures, a classic problem in engineering design [24], aims to determine a configuration of nodes and members such that the resulting structure can sustain an arbitrary load. Further, a sequential decomposition allows the designers to modify the design state incrementally, building the design step by step. A human design study was previously conducted to record sequential human actions while solving this problem [68]. The study comprised 48 humans working in teams of 3 to create a truss design with sufficient strength (quantified as Factor Of Safety, FOS) and minimal mass. This dataset has been used in prior work to train Deep Learning Agents (DLAgents) and their variants [25,34,66] and has also been used to train probabilistic models to explore action sequencing [18,69]. The training of the proposed DSN framework only considers the generative actions that positively add components or increase their sizes (i.e., Add a Node, Add a Member, and Increase Thickness of a member) and subtractive actions that remove elements or decrease size are ignored. Even though this leads to a significant reduction in the dataset size, it focuses on actions that lead to *building* trusses (rather than revising or trimming them), thereby reducing the noise in the dataset. Each of the mentioned three actions has a discrete component (the action label) and a spatial component called parameters $(x_1, y_1, x_2, y_2)$ or $(x_1, y_1)$ that define the particular action as specified in Table 1. It can be observed that Add a Member and Increase Thickness are state-dependent discrete actions since their domains are constrained to the existing set of nodes and members, respectively. This domain also expands as the set of nodes and members increases over time. A rough estimate[1] of the number of possible design trajectories that can exist for this problem is of the order of $10^{1000}$, which is much larger than other complex problems ($10^{123}$ for chess and $10^{180}$ for Go) [8].

*Table 1:* Description of truss design actions

| Discrete Action (Label) | Parameters | Domain | Type |
|---|---|---|---|
| Add a Node | x, y | $x, y \in$ Continuous 2D space | Continuous |
| Add a Member | $x_1, y_1, x_2, y_2$ | $(x_1, y_1, x_2, y_2) \in$ Node Set | State-dependent |
| Increase Thickness | $x_1, y_1, x_2, y_2$ | $(x_1, y_1, x_2, y_2) \in$ Member Set | State-dependent |

The dataset contains a total of 8048 samples with information about the design state as an image and the design action selected by the human designer. Truss design actions require at most two spatial control points with x-y coordinates for the two endpoints of a member, so the spatial region is defined by a mixture of two bivariate isotropic Gaussian distributions ($S = 2$). Preprocessing the data requires creating the two datasets $D_1 = \{s_t ; \ p_t^s\}$ and $D_2 = \{(s_t, (p_t^d, p_t^s)^A); i\}$. The human dataset already contains the state information ($s_t$) and the finally selected action information ($p_t^d, p_t^s$); however, the feasible set of actions $A\_t \, | s_t, p_t^s = (p_t^d, p_t^s)^A$ needs to be computed for the given state and predicted spatial region. For training, the spatial region is centered around the selected action's exact spatial parameters $p_t^s$; hence, the set of feasible actions is sampled from the action's spatial neighborhood. In the case of truss design problem $p_t^d$ is a 3-dimensional one-hot vector since there are three action labels and $p_t^s$ is a 4-dimensional vector that contains the spatial parameters. In the case of *Add a Node* the spatial parameters are padded to maintain a uniform vector length of 4 across all action parameters. The total number of feasible actions in a state is limited to 50 ($A_{max}$), including 10 ($n$) continuous actions per state. These numbers are empirically selected and may be further optimized given the goal of the learning process, for which a more rigorous and computationally intensive study may be required. This limitation allows the spatial region to filter out the top 50 feasible actions in each state based on the spatial region distribution likelihood.

## 4.2 Network details

This subsection details the three deep learning networks used in the DSN framework developed for the truss design problem. The three different networks are named: Encoder, Spatial Action, and Selection.

*Encoder Network:* Figure 3 provides the schematic for this network. The encoder follows an architecture similar to the Atari DQN network [70]. There are three convolutional layers with decreasing kernel sizes ($8\times8$, $4\times4$, and $3\times3$) followed by a linear layer that generates the 512-dimensional final latent representation. Every layer uses a ReLU non-linearity. In order to reduce the number of dimensions on the image data, large strides are used [71]. The number of channel features progress from 3 in the input to 32, 64, and 64 in the three consecutive convolutional layers. The encoder takes the state input as an RGB 3 channel $128\times128$ pixel-based input and generates a 512-unit latent encoding.

*Spatial Network:* The spatial network employs a basic 3-layer neural network with 64, 32, and 4 units, respectively, as shown in Figure 4. The first two layers of the network have ReLU activation, and the final layer has TanH activation to bound the output between -1 and 1 (the bounds of design space). The final layer has four units since there are four spatial parameters required for truss design, as shown in Table 1 however, it can be arbitrarily modified to suit other problems.

*Selection Network:* The selection network architecture aims to learn the relationship between the combined state and feasible action set with the final selected action. The network must handle an unordered set of inputs with variable sizes as the feasible set of actions keeps changing given the state. The idea is to generate a combined state-action representation for each action in the set. It follows an invariant style network similar to a PointNet architecture [36], where it transforms each item in the set by several layers and then uses an aggregating function at the end; the schematic for the network is shown in Figure 5. Every action's representation comprises both discrete and spatial components. For each action, the spatial parameters are transformed with four layers of progression (64, 128, 256, and 256). The discrete parameters (action label) only go through one layer of transformation from 3 to 64 and are then concatenated together into a vector (320 units), then transformed into 512 units. A concatenation operation then joins each of the actions with the latent encoding of the encoder network's state. As a result, a joint representation for every state and action pair is generated. Each pair representation is further transformed and downsampled from 512 to 128 and then finally to 32 units. Each layer in the transformation of these action representations has a Batch Norm followed by a ReLU activation to enable proper learning. The data at this point is sized $N \times 32$; an average pool with a 32 sized kernel brings it further down to $N \times 1$, generating a scalar value for every action. Finally, this culminates with a SoftMax aggregation that produces a probability distribution across all actions in the set. Since SoftMax is an order invariant aggregator, it helps the network learn effectively from the unordered set of action inputs. It must be noted that the selection network is independent of the size of the action input and can deal with a set of $N$ action inputs with variable size.

---

[1] This estimate discretizes the continuous design space into a 0.1-unit grid, which corresponds to about $10^4$ continuous actions and $2 \times^n C_2$ discrete state-dependent actions per iteration where $n$ is the number of existing nodes in the design state. On average the design process had 250 iterations in the human study [68]. Hence the overall problem space tree is: $10^{4^{250}} + (2 \times^n C_2)^{250} \geq 10^{1000}$.

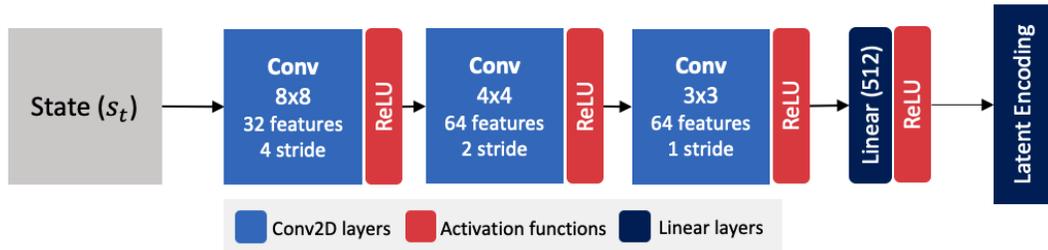

*Figure 3:* Schematic representation of the Encoder network layers

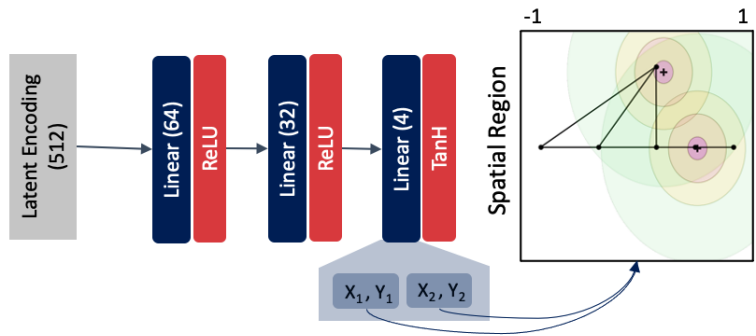

*Figure 4:* Schematic of the Spatial Action network

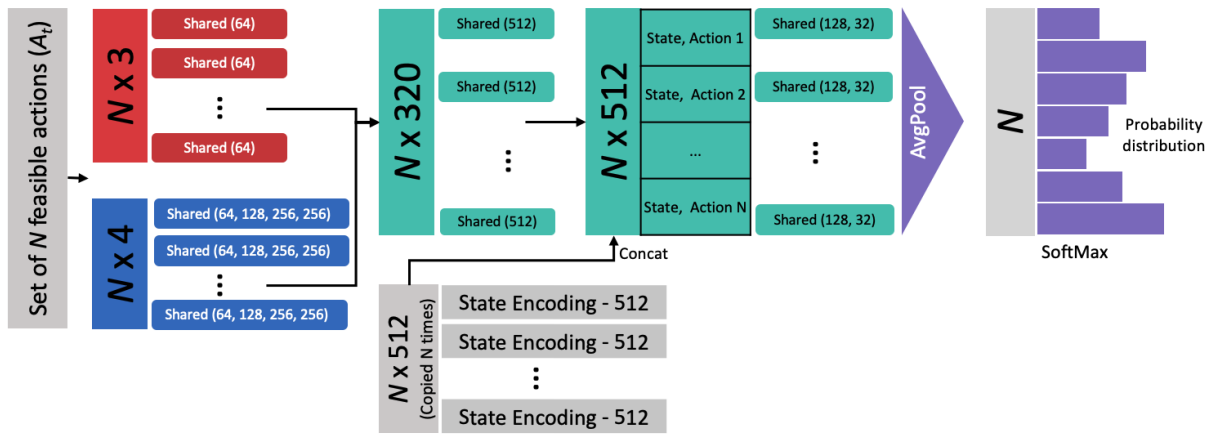

*Figure 5:* Schematic of the Selection network. Shared refers to having the same weights across all $N$ inputs leading to identical transformations across the $N$ actions.

**4.3 Experimental Setup**

To effectively train and evaluate the framework, 10-fold cross validation [72] is carried out. The dataset is split into 10 sections (folds) and one of them is used for testing while other 9 are used for training so that the testing set comprises of unseen design states. Further, 10% of the training data in every fold is used as a validation set. The training process is then repeated 10 times where each fold is used as a testing set once. This is a comprehensive method to evaluate the model performance as it normalizes any variance in the accuracies due to the test-train split. For each iteration the DSN framework is trained using the Adam optimizer [65], and a decaying learning rate for 200 epochs starting from 0.001 with early stopping if no improvement is seen over 20 epochs. The best performing network in terms of accuracy (Top-1 selection accuracy) is identified and stored. The two different loss values (MSE and BCE) are observed and compared. These loss values provide an aggregated value of error over the dataset that helps to compare the network's performance. Even though these losses are adequate for training, they only provide a one-dimensional view of the performance. Two accuracy metrics are devised to further analyze and evaluate the error distribution over the whole test set.

The two prediction accuracy metrics are explicitly developed to evaluate the performance of the two hierarchies of the framework: spatial accuracy and selection accuracy. Spatial accuracy evaluates if the predicted spatial region is within the error threshold bounds of the ground truth spatial region. For example, a threshold value of 0.3 units means that both the control points of the spatial region must lie within 0.3 distance units from the ground truth control points to be classified as a correct prediction. The distance units are normalized to represent the entire design space within the range of [-1, 1] (two extremes of the image); hence a tolerance of 0.3 distance unit is equivalent to about 15% of the overall design space length. The threshold value can be changed to observe the performance with strict or lenient error tolerances. Increasing the threshold value makes it easier for the framework to achieve higher accuracies. Selection accuracy represents how well the framework can predict the final action selected by the human. The accuracy is defined as Top-k selection accuracy, which classifies a prediction to be accurate if the top k predictions from the model contain the ground truth action selected by the human. Here Top-1 prediction is analogous to a standard accuracy metric where only the best action is considered. Like spatial accuracy, increasing k (or error threshold for successful prediction) makes it easier to achieve high accuracy values. These threshold values allow evaluation of the performance variation across the whole test set dataset providing more granular information than losses. It must be noted that these prediction accuracy metrics have only been utilized for analyzing the test performance and have not been used for training the networks.

Although DSN is a unique framework that hierarchically decomposes action prediction allowing compatibility with complex actions, an Atari-DQN based model was trained to benchmark. The original network outperformed humans on multiple Atari games and illustrates the effectiveness of such policy networks on discrete action space problems. The model consists of three convolutional layers similar to the original paper [42] that reduce the dimensions of the image from $128 \times 128 \times 3$ to $12 \times 12$, followed by two linear layers to predict both discrete and continuous parameters of the hybrid action definition (similar to [37]). This network is trained similarly to DSNs, but instead of predicting the final action from a set of feasible actions, it predicts the discrete and continuous parameters of the final actions directly without any information about the feasibility of the actions. It uses the original version of the dataset $D = \{s_t; (p_t^d, p_t^s)\}$ $\forall$ t = [1, T], with T data samples, that has the direct relationship between the state and the final selected action. This network is referred to as the "Imitation Network" since it is a vanilla non-hierarchical policy network that directly learns to imitate human action's spatial and discrete parameters. This model respects the hybrid nature of the action space (discrete and continuous parameters); however, it does not comply with the state-based constraints and hence can make predictions in infeasible action spaces. To overcome this, an approximation is used; the predicted spatial parameters are used to find the nearest feasible action from the sampled set based on Euclidean distance. This approximation allows the network to be applicable for complex action spaces and predict (the nearest) feasible action and spatial regions for a baseline comparison in a complex action space problem. It must be noted that both DSN and Imitation Network use the spatial region-based feasible action sampling described in Section 3.2 with the primary difference being, DSN utilizes the action set as an input since it can deal with unordered variable set-based input and generate a probability distribution while the Atari-DQN uses a Euclidean distance heuristic estimate to predict the final action selection from the set. The number of parameters for DSN is slightly higher than the non-hierarchical Imitation Network (5.17 million parameters versus 4.8 million parameters) since the additional network (selection network) is employed to generate the final action probability distribution.

 

## 5. RESULTS AND DISCUSSION

This section presents the validation results from the prediction accuracy comparison of the two networks, followed by a qualitative discussion over the usefulness and failure cases of the proposed network.

### 5.1 Evaluating framework prediction accuracy

Comparing the average loss metrics and average prediction accuracies help to determine how well a model can predict human decision-making behavior from historical data. These metrics are compared for two different networks: the proposed Design Strategy Network (DSN) and the vanilla Imitation Network from Atari-DQN. Table 2 shows the different metrics averaged over 10 runs of 10-fold-cross validation for comparison. The MSE loss shows the error for predicting the spatial region. The DSN framework (0.0215) outperforms the Imitation Network (0.0252) by 15j%, showing its ability to predict the preferred spatial region more accurately. It must be noted that even though both DSN and Imitation Network use very similar networks (architecture and layers) to predict the spatial regions, DSN perform much better in terms of Spatial accuracy and MSE loss values. In terms of prediction accuracy, only the metrics with the tightest thresholds are shown in Table 2. Spatial accuracy (0.1 units) illustrates the percentage of test data the network can predict within a 0.1 error tolerance. It can be observed that the proposed DSN framework performs better than the Imitation Network baseline (56% vs. 31%) with a 0.1 threshold. Similarly, selection accuracy depicts the percentage of test data for which the network exactly predicts the final action. The DSN outperforms the Imitation Network on selection accuracy (74% vs. 48%) for the Top-1 action prediction. The proposed DSN framework significantly outperforms the existing method for policy representation and behavior prediction in all the metrics. To further analyze this performance, the accuracy threshold values are varied.

*Table 2 Comparing the testing performance of the DSN framework with respect to the vanilla Imitation Network over 10-fold validation. The values are averaged over 10 runs and show ± first standard error.*

| Model Type | MSE loss | BCE loss | Spatial Accuracy (0.1 units) | Selection Accuracy (Top 1) |
|---|---|---|---|---|
| Imitation Network | 0.0252±0.0015 | 0.1524±0.0055[2] | 31.31±1.88 | 48.00±1.50[3] |
| DSN | **0.0215±0.0011** | **0.0326±0.0009** | **56.17±1.79** | **73.82±00.94** |

Different accuracy thresholds for spatial accuracy and selection accuracy are compared using stacked horizontal bar graphs in Figures 6 and 7. Here, the x-axis represents the average percentage of the test set going from 0-100. The bars from left to right represent the different error threshold levels, and the width of the block represents what percentage of the test set that threshold correctly predicts. These percentage values are also displayed from left to right (minimum to the maximum threshold) and explicitly represent the width of the block corresponding to the percentage of the test set. This stacked bar graph visualizes the discrete distribution of the test set across the error thresholds. It provides more granular information about the performance of the frameworks than aggregated MSE and BCE values, which may help compare different frameworks depending on the application. Further, this can isolate outliers and provide more interpretable information about the performance.

Figure 6 shows the average testing spatial accuracy distribution over 10 runs for the two frameworks. The learning problem essentially aims to predict human visual attention, meaning the area of the design space the designer is focusing on, accurately predicting which is a challenging problem. Four different tolerance values are evaluated: 0.1, 0.3, 0.5, and 1.0. The graph for DSN can be read as follows; 56.17% of test data lies within 0.1 error of prediction, 30.45% lie between (0.1, 0.3] error, 7.64% with (0.3, 0.5] error, 4.74 % have an error between (0.5, 1.0] and 0.73% have an error higher than 1.0 units. Benchmarking performance indicates the Imitation Network does decently; however, it cannot match the DSN framework's performance with the additional Selection network for making the final action selection. In comparison, a random prediction is close to 0% for low tolerance of 0.1 units as it is challenging to predict this accidentally. Over 60% of data cannot be predicted with random guessing even with >1.0-unit tolerance; the bar graphs for the random case illustrate the difficulty of the learning problem.

---

[2] The BCE losses cannot be directly compared for the two frameworks. The Imitation Network only predicts the discrete component of the final action and hence only has three classes compared to 50 classes for DSN (Selection Network), which predicts the final action from the feasible set.

[3] To compute the selection accuracy the nearest feasible action is compared with the ground truth, an approximation required for Imitation Network to output feasible actions.

Figure 7 shows a comparison of Top-k selection accuracies for the two models. Four values of k for Top-k accuracy are compared: 1, 3, 5, and 10. Here Top-1 prediction is analogous to the typical accuracy where selecting the exact action will lead to a correct prediction. A considerable difference in Top-1 selection accuracy is observed between the DSN framework (74%) and Imitation Network (48%), again illustrating the effectiveness of the hierarchical nature of the framework. The bars from left to right represent if the correct human action is amongst the 2nd and 3rd prediction, 4th and 5th prediction, 6th to 10th prediction, or more than the top 10 predictions. Selection accuracy (especially Top-1) provides a stringent metric to compare if the framework can precisely predict the human designer's action, despite that DSN can achieve an accuracy value of 74%, which illustrates the effectiveness of the framework to predict human decision-making behavior on unseen truss states. The random model bar plots further illustrate the difficulty of the challenge, and a significant gap between the random prediction and the DSN is observed. These frameworks successfully extract and represent these relationships in their deep networks and then leverage them to predict human behavior in unseen design states.

The results illustrate the effectiveness of DSN at predicting human behavior. It predicted human behavior in unseen design states with accuracies much higher than the baseline Imitation Network and a random baseline. This effectiveness implies that using this hierarchical framework can capture human intuition including their underlying domain knowledge, from historical data to simulate that behavior.

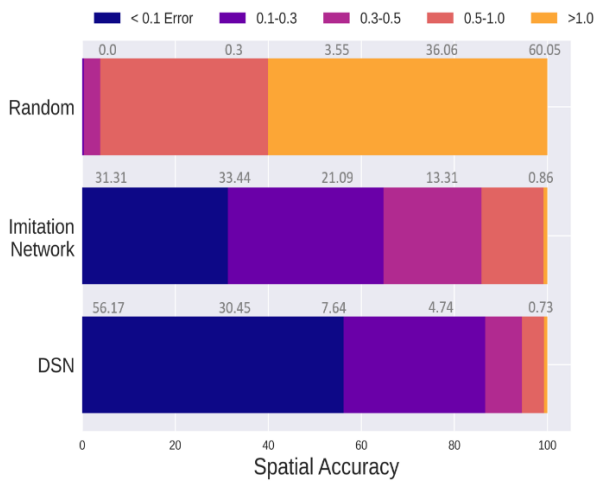

*Figure 6:* Spatial accuracy distribution over the test set with % data cover for given error thresholds.

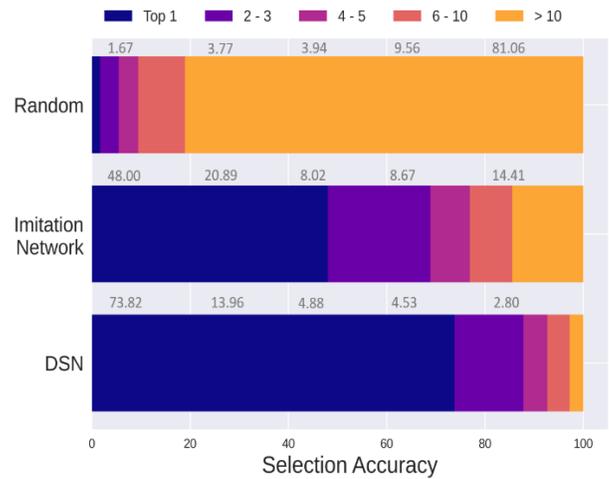

*Figure 7:* Selection accuracy distribution over the test set with % data cover for given error thresholds in Top-k accuracy.

## 5.2 Qualitative evaluation

Further qualitative evaluation is conducted by visualizing and comparing the predicted spatial region and final actions with the ground truth values selected by the human from one of the testing sets of 10-fold-validation. Figure 8 illustrates spatial region prediction for four different samples. The spatial region is centered at the '+' symbol and consists of 4 concentric circles. The circles represent the 0.1 (smallest), 0.3, 0.5, and 1.0 (largest) tolerance values used to evaluate the spatial accuracy. It can be observed that the innermost circle, which corresponds to 0.1 tolerance, refers to a tiny area relative to the design space. It can be observed that DSN can predict the spatial region very close to the ground truth, performing much better than the Imitation Network. The difference is particularly high in examples 3 and 4, where the Imitation Network predicts some part of the region correctly but cannot attain a sub 0.1 tolerance value for both the points. Example 3 illustrates a continuous action (addition of node) corresponding to overlapping spatial regions since there is only one control point; this is accurately predicted by the DSN framework showing its superiority in performance. It appears that the Imitation Network can predict the general region of action, which is close to the ground truth, meaning it has captured a rough idea of the visual attention model of humans; however, it is not highly accurate. The third and fourth examples illustrate the inaccuracies where DSN accurately made the predictions.

Figure 9 compares the two frameworks for the final action prediction using the same four examples. These four examples have been selected to illustrate the complexity of the problem since multiple potentially good actions exist in these states, and the network must predict the one selected by the human designer in the unseen dataset. The four columns represent the input design state, ground truth, predicted action by the DSN framework, and the benchmark framework. The action is highlighted using shaded rectangles in these figures. It can be observed that in the first two examples, both DSN and Imitation Network can correctly predict the action; however, when the design becomes more complex (with numerous action options), as shown in examples 3 and 4, Imitation Network begins to make errors in the prediction. Here again, it can be observed that having the proposed hierarchical decomposition in strategy representation can achieve high prediction performance

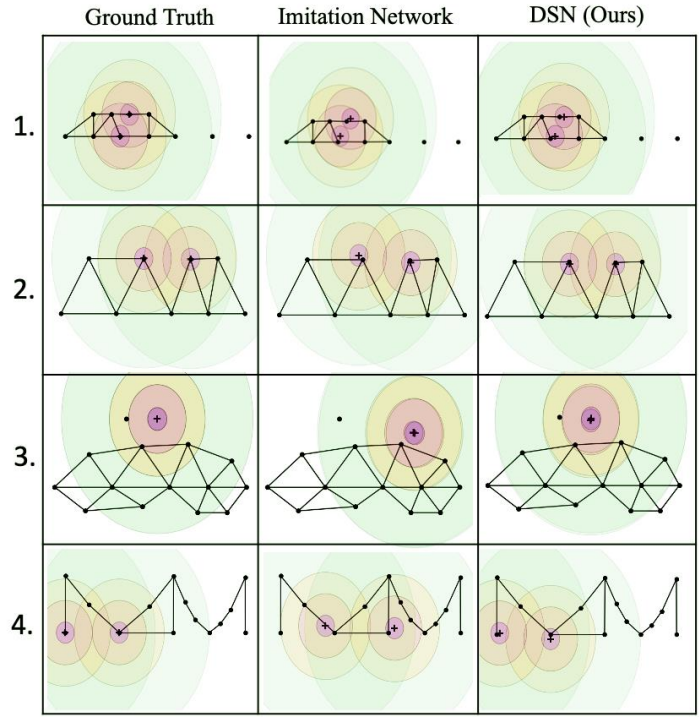

*Figure 8:* Comparative evaluation of DSN and Imitation Network frameworks for spatial region prediction. Shaded regions centered around '+' are the two control points of the spatial regions. The concentric circles represent tolerance values of 0.1, 0.3, 0.5 and 1.0.

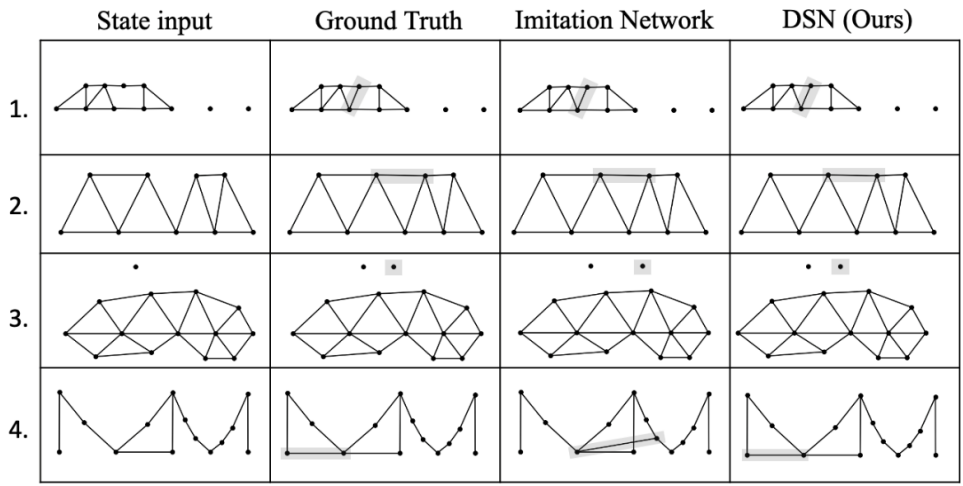

*Figure 9:* Comparative evaluation of DSN and Imitation Network frameworks for final action prediction on the test dataset.

 

Extending the qualitative evaluation to failure examples from the data provides exciting insights into the performance of these networks. The absolute metric that this work utilizes is the prediction accuracy of the networks. However, predicting human actions is a very challenging task, and producing a 100% accurate model of humans, in general, may not be possible or even beneficial. This work aims to achieve as high accuracy as possible to allow the network to capture the general intuition of decision-making in design. Figure 10 illustrates two examples that show the current state, spatial region, ground truth, and the final action prediction by the high-performing DSN framework. In the first example, the network predicts a spatial region in the vicinity of the actual action; however, while predicting the final action, it identifies that "Add a Member" needs to be selected; however, the spatial location is not precisely what the human selected leading to an incorrect action selection. Similarly, in the second example, the network correctly predicts "Increase Thickness"; however, the spatial location is slightly different from the original human action and is again classified as an incorrect selection. It must be noted that the final action is selected from a set of feasible actions and hence must be identical to the ground truth for accurate classification. The ability of the network to discern this level of detail in the design problem is one of the significant contributions of this DSN framework. The hierarchical network can generate this detail due to the decomposition of the prediction problem into determining a spatial region first and then selecting from the sampled list of actions. Moreover, it is not just the top-1 selection accuracy of 74% that is valuable; instead, if the top-5 actions are observed, the prediction firstly contains the selected human action (with 93% accuracy) and also contains several other 'inspired' actions that if explored further may turn out to lead to better designs and hence also very valuable. This multiple action output enables the framework's application in search-based optimizers and the development of design agents to simulate high-performing design decision-making and identify potentially meaningful search directions. As a result, this framework can guide synthetic design generation problems and even integrate auxiliary objectives to enhance the design generation performance by using this learning as an initial "human prior". An important challenge when using an imitation learning framework for a generative task is Distribution Shift [74,75]. Distribution shift refers to an instance when the network makes a prediction on a state that is largely different from the training state distribution, leading to an incorrect or seemingly random prediction despite good performance during training. This is because during training the network learns only expert strategies and hence does not learn how to deal with or recover from non-expert design states. This is a standard problem with all imitation learning approaches and limits the performance of generated trajectories in novel domains. As a solution, these imitation learning methods are often bootstrapped with search [25,74] and optimization methods (often reinforcement learning based [39,75,76]) that lead to enhanced performance and effective utilization of the strategies learned by the policy networks.

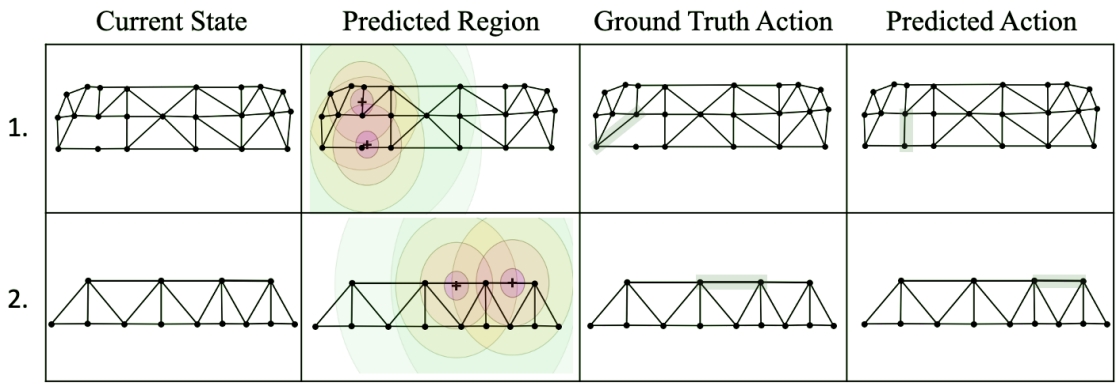

*Figure 10:* Divergence between the DSN framework and the actual human prediction at highest accuracy

## 6. CONCLUSION

Extracting and representing design decision-making strategies from behavioral data is one of the fundamental challenges in data-driven design research. Design domains introduce several challenges in this area involving the representational complexity of action spaces and the diversity in the design state information. This work presents a novel hierarchical framework architecture for policy or strategy learning in design termed Design Strategy Network (DSN). The framework aims to represent decision-making

 

mechanisms in the domain of design problems, more specifically sequentially generative configuration design problems with arbitrary decision spaces. The framework is designed to predict actions given a particular design state. It leverages the concept of visual attention to break down the prediction process into first identifying a rough spatial region in the design space and then selecting the final action in that neighborhood. The DSN framework in this work is trained on a dataset of many humans solving a truss design problem. The framework learns to predict the human action throughout the design sequence, including all the details and information required for the comprehensive action definition and achieves an overall accuracy of 74%. Throughout the experiments in this paper, the framework illustrates a successful implementation where the network can navigate the state-action relationships over a divergent hybrid action space.

The experiments detailed in the paper address the main research question of the work: *Can data-driven learning with hierarchical actions emulate human decision-making behavior in complex action spaces better than a baseline non-hierarchical method?*

The DSN framework uses a combination of Encoder network, Spatial Action network, and Selection network. It first generates a latent encoding of the image-based design state-input, which the Spatial Action network transforms to predict a preferred spatial region of the final action. The design environment is then used to generate a set of actions in that spatial region. Finally, the Selection network generates a probability distribution over this set of actions to predict the final action. This ability of the selection network to input an arbitrarily long set of feasible actions allows the framework to be applied for any arbitrary action space since the formulation does not depend on its size or complexity. After training, the network can achieve higher accuracy values than baseline methods and predict both the spatial region and the final action selection. Since the framework can effectively predict human behavior on unseen design states, the framework has captured certain essential components of design knowledge, mainly sequential generative rules that implicitly govern human decision-making behavior.

The significant contribution of this work is the generic deep learning-based model that can effectively capture relationships between state and design decisions for an arbitrary action-space sequentially-generative design problem. The framework is potentially adaptable for problems with a state representation of the form of an $N$-dimensional matrix. The different components of this framework learn everything from data and can be applied across a problem with minimal changes to the architecture. This work goes beyond previous works in imitation learning by using a variable set of actions as network input. The deep learning network explicitly learns state-action relationships over an arbitrarily defined complex action space individually evaluating all state and feasible action pairs without the need of any rule-based inference algorithms to guide decision making [34]. Even though the network details are specific to the truss design problem, the formulation of the learning problem of the three independent networks is independent of the truss design problem. This allows the framework to be generically applied to other similarly modeled sequentially generative design problems; however, the final performance may require hyperparameter optimization to reach the optimal solutions.

## ACKNOWLEDGEMENTS

This material is based upon work supported by the Defense Advanced Research Projects Agency through cooperative agreement No. N66001-17-1-4064. Any opinions, findings, and conclusions or recommendations expressed in this paper are those of the authors and do not necessarily reflect the views of the sponsors.